\documentclass{article}
\usepackage{spconf,amsmath,graphicx}

\title{Effect of Lossy Compression Algorithms\\ on Face Image Quality and Recognition}
\name{Torsten Schlett \qquad Sebastian Schachner \qquad Christian Rathgeb \qquad Juan Tapia \qquad Christoph Busch
\thanks{%
This work was funded by the German Federal Ministry of Education and Research and the Hessian Ministry of Higher Education, Research, Science and the Arts within their joint support of the National Research Center for Applied Cybersecurity ATHENE.
This project has also received funding from the European Union’s Horizon 2020 research and innovation programme under grant agreement No 883356.
This text reflects only the author’s views and the Commission is not liable for any use that may be made of the information contained therein.
Portions of the research in this paper use the FERET database of facial images
collected under the FERET program, sponsored by the DOD Counterdrug Technology
Development Program Office \cite{Phillips-FERET-1998}\cite{Phillips-FERETEvaluationMethodologyFaceRecognition-PAMI-2000}.
}
}
\address{Hochschule Darmstadt, Germany}

\usepackage{hyperref}
\usepackage{multirow}
\usepackage[export]{adjustbox}

\graphicspath{{./images/}}

\newcommand{\eg}{e.g.\@}
\newcommand{\etal}{et al.\@}

\begin{document}
\nocite{ISO-IEC-2382-37-2022}\nocite{ISO-IEC-39794-5-2019}\nocite{Funk-QualityPerformance-IEEE-WestPoint-2005}
\maketitle
\begin{abstract}
Lossy face image compression can degrade the image quality and the utility for the purpose of face recognition.
This work investigates the effect of lossy image compression on a state-of-the-art face recognition model, and on multiple face image quality assessment models.
The analysis is conducted over a range of specific image target sizes.
Four compression types are considered, namely JPEG, JPEG 2000, downscaled PNG, and notably the new JPEG XL format.
Frontal color images from the ColorFERET database were used in a Region Of Interest (ROI) variant and a portrait variant.
We primarily conclude that JPEG XL allows for superior mean and worst case face recognition performance especially at lower target sizes,
below approximately 5kB for the ROI variant,
while there appears to be no critical advantage among the compression types at higher target sizes.
Quality assessments from modern models correlate well overall with the compression effect on face recognition performance.
\end{abstract}

\section{Introduction}

Image quality in terms of biometric utility \cite{ISO-IEC-2382-37-2022} influences the performance of face recognition systems,
as \eg{} used for automated border control.
Lossy image compression can affect the utility,
and is relevant \eg{} for Machine Readable Travel Documents (MRTDs) \cite{ISO-IEC-39794-5-2019}.
Since the effect on utility may differ substantially between lossy compression schemes at approximately the same compressed image size,
this leads to the question which compression type should be preferred.
Furthermore, there are automatic Face Image Quality Assessment (FIQA) methods that attempt to estimate an individual image's biometric utility,
leading to the question whether these methods' assessments are similarly influenced by image compression in isolation.

\begin{figure}
\centering
\includegraphics[width=\linewidth]{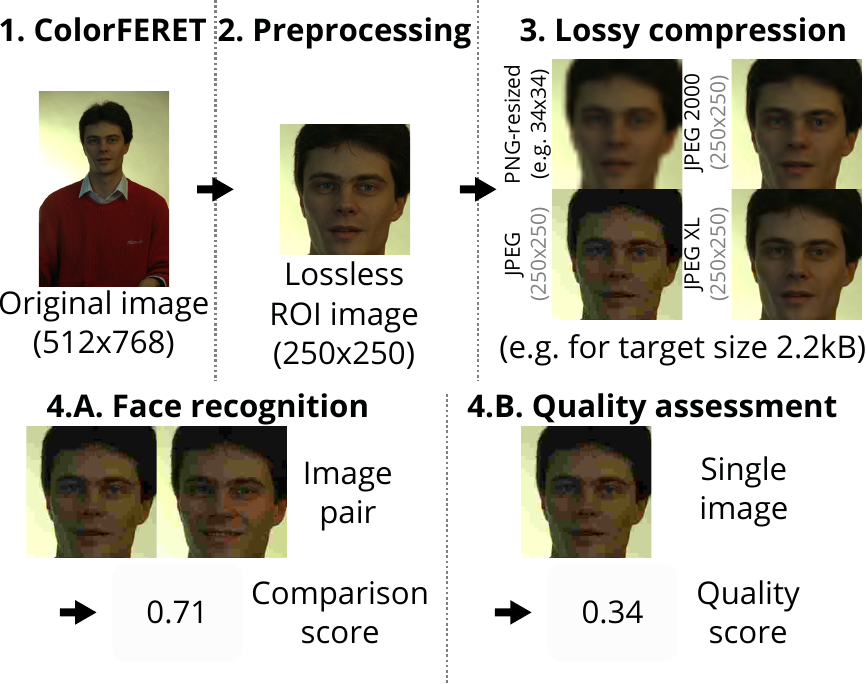}
\\\vspace{-1em}
\caption{\label{fig:overview} Processing stages required for the experiments.}
\vspace{-2em}
\end{figure}

As an example for related prior work,
Funk \etal{} \cite{Funk-QualityPerformance-IEEE-WestPoint-2005} evaluated JPEG and JPEG 2000 with respect to two unspecified face recognition systems on 358 frontal face images from the FERET \cite{Phillips-FERET-1998}\cite{Phillips-FERETEvaluationMethodologyFaceRecognition-PAMI-2000} database, all $512\times 768$ pixels, finding that JPEG 2000 outperformed JPEG below a size of 17kB.
Later work by Delac \etal{} \cite{Delac-ImageCompressionEffectsFaceRecognition-2007}
likewise evaluated JPEG and JPEG 2000 on images from FERET,
but for different grayscale subsets transformed and cropped to a $128\times 128$ pixels ROI with masked background,
for three face recognition algorithms (based on Principal Component Analysis, Linear Discriminant Analysis, Independent Component Analysis),
similarly finding JPEG 2000 results superior to JPEG for higher compression rates.
Our work contributes an evaluation of the effect of lossy compression on face recognition using a modern ArcFace \cite{Deng-ArcFace-IEEE-CVPR-2019} model,
comparing the new JPEG XL \cite{ISO-IEC-18181-1-2022} encoding
besides the previously evaluated JPEG \cite{ISO-IEC-10918-1-1994} and JPEG 2000 \cite{ISO-IEC-15444-1-2019},
as well as downscaled PNG \cite{ISO-IEC-15948-2004} images,
which proved to be competitive with the dedicated lossy compression methods.
Our evaluation makes use of frontal color images from ColorFERET \cite{Phillips-FERET-1998}\cite{Phillips-FERETEvaluationMethodologyFaceRecognition-PAMI-2000}
across a range of concrete image target sizes.
Additionally,
we examined how well the effect of lossy compression on face recognition is reflected in the per-image quality scores produced by eight different FIQA methods,
which previously have only rarely been examined in a lossy image compression context \cite{Schlett-FIQA-LiteratureSurvey-CSUR-2021}.
\autoref{fig:overview} provides an overview of the processes required for the experiment data.

In the remainder of this paper, \autoref{sec:setup} describes the details of the experiment setup, \autoref{sec:results} presents an analysis of the results, and \autoref{sec:conclusion} concludes with the main findings.

\vspace{-1em}\section{Setup}\vspace{-1em}
\label{sec:setup}

\subsection{Database}\vspace{-0.2em}

A high initial quality of the images is desired for the experiments,
to better isolate the impact of lossy image compression for the analysis.
Focusing on frontal images further allows us to approximate a passport image scenario.
Thus the frontal color images from the ColorFERET \cite{Phillips-FERET-1998}\cite{Phillips-FERETEvaluationMethodologyFaceRecognition-PAMI-2000} database were used for the experiments.
This amounts to 2638 images across 966 subjects.
These images were captured under comparatively controlled conditions,
in contrast to \eg{} a web-scraped face image database such as LFW \cite{LFWTech}.
The facial expression does however vary (neutral or smiling),
and for some subjects there are images both with and without glasses.

Many of the original ColorFERET \cite{Phillips-FERET-1998}\cite{Phillips-FERETEvaluationMethodologyFaceRecognition-PAMI-2000} images show a larger area than just the face,
so we applied preprocessing to obtain ROI images for the experiments,
as illustrated in \autoref{fig:overview}.
Images were cropped and aligned as required for direct usage by the ArcFace \cite{Deng-ArcFace-IEEE-CVPR-2019} face recognition model,
except for the resolution, which was $250\times 250$.
The landmark detection required for preprocessing was done via the publicly available RetinaFace-R50 model \cite{Deng-FaceDetection-RetinaFace-CVPR-2020} from the InsightFace project.
Both the face recognition model and the six deep learning FIQA models require various fixed input image sizes,
all smaller than $250\times 250$,
thus images were resized with bilinear interpolation for each of these models.

\vspace{-1em}\subsection{Face recognition}\vspace{-0.2em}

To obtain comparison scores \cite{ISO-IEC-2382-37-2022} we used the publicly available ArcFace-R100-MS1MV2 \cite{Deng-ArcFace-IEEE-CVPR-2019} face recognition model from the InsightFace project,
which requires a $112\times 112$ input image size.
Multiple comparison trial \cite{ISO-IEC-2382-37-2022} configurations were analyzed:
The ``mated-other'' configuration consisted out of all 3484 possible mated \cite{ISO-IEC-2382-37-2022} trials for the subjects,
and compared different captured images from the same subject with the same target size compression applied.
The ``mated-self'' configuration on the other hand consisted out of comparisons of the 2638 images with lossless compression
against the same face image (stemming from the same captured sample) with lossy compression applied.

\vspace{-1em}\subsection{Quality assessment}\vspace{-0.2em}

To obtain quality scores we used eight FIQA methods.
For all of the methods, higher quality score output is intended to indicate higher biometric utility \cite{ISO-IEC-2382-37-2022}.
Six of these were modern deep learning models specifically intended to assess quality in terms of image utility \cite{ISO-IEC-2382-37-2022} for face recognition:
CR-FIQA(S) \& CR-FIQA(L) \cite{Boutros-CRFIQA-arXiv-2021}
(respectively with iResNet50/iResNet100 backbone trained on CASIA-WebFace\cite{Yi-LearningFaceRepresentationFromScratchCASIAWebFace-arXiv-2014}/MS1MV2\cite{Deng-ArcFace-IEEE-CVPR-2019}, both with $112\times 112$ input image size),
MagFace \cite{Meng-FRwithFQA-MagFace-CVPR-2021}
(iResNet100 backbone trained on MS1MV2 \cite{Deng-ArcFace-IEEE-CVPR-2019}, $112\times 112$ input image size),
SER-FIQ \cite{Terhorst-FQA-SERFIQ-CVPR-2020}
(``same model'' variant using ArcFace, $112\times 112$ input image size),
FaceQnet-v0 \cite{Hernandezortega-FQA-FaceQnetV0-ICB-2019} \& FaceQnet-v1 \cite{Hernandezortega-FQA-FaceQnetV1-2020}
(both with ResNet-50 backbone, trained on VGGFace2 \cite{Cao-VGGFace2Dataset-FGR-2018}, $224\times 224$ input image size).
These six models are all publicly available.
The other two FIQA methods were handcrafted sharpness measures,
referred to as Sharpness-1 and -2:
Sharpness-1 is implemented as described by \cite{Crete-PerceptualBlurMetric-2007}, comparing the pixel intensity variations in the input image against those in an artificially blurred version thereof to estimate the sharpness.
Sharpness-2 is conceptually similar but simpler, computing the sharpness as the mean of the distances between the normalized $[0, 1]$ pixel values of the input image and a blurred ($3\times 3$ mean box filter) version thereof.

\vspace{-1em}\subsection{Compression target sizes and types}\vspace{-0.2em}

Target sizes for the lossy image compression were selected manually,
based on the lowest usable size among all used compression types,
and based on the sizes of the images as lossless PNG files.
Images were compressed as closely as possible to each target size, with the target size serving as the inclusive upper limit, meaning that an image's size may at most be equal to the target size.
We selected the target sizes 5kB, 4.5kB, 4kB, 3.5kB, 3kB, 2.5kB, and 2.2kB.

Four lossy compression types were examined:
PNG-resized, JPEG \cite{ISO-IEC-10918-1-1994}, JPEG 2000 \cite{ISO-IEC-15444-1-2019}, and JPEG XL \cite{ISO-IEC-18181-1-2022}.
PNG-resized refers to PNG \cite{ISO-IEC-15948-2004} images that were downscaled to reach each target size,
keeping the aspect ratio of the source image as closely as possible,
thus effectively turning the by itself lossless PNG compression into a lossy compression scheme.
PNG, JPEG and JPEG 2000 are permissible as face image encodings in MRTDs according to ISO/IEC 39794-5:2019 \cite{ISO-IEC-39794-5-2019}.
JPEG XL is a more recently created and standardized encoding.
For PNG, JPEG, and JPEG 2000 compression we used Pillow version 9.1.0,
which in turn relies on ZLIB version 1.2.12 for PNG, on libjpeg-turbo version 2.1.3 for JPEG, and on OpenJPEG version 2.4.0 for JPEG 2000.
For JPEG XL we used the reference implementation libjxl version 0.6.1.
Default compression settings were used for all compression types, except for the comparatively unusual PNG-resized compression, for which we did set the Pillow ``optimize'' option.

Step three in \autoref{fig:overview} shows examples for ROI images compressed with all four types at target size 2.2kB.

\section{Results}
\label{sec:results}

\vspace{-1em}\subsection{Effect on comparison scores}\vspace{-0.2em}

\begin{figure}
Mated-other:\\
\includegraphics[width=\linewidth,valign=c]{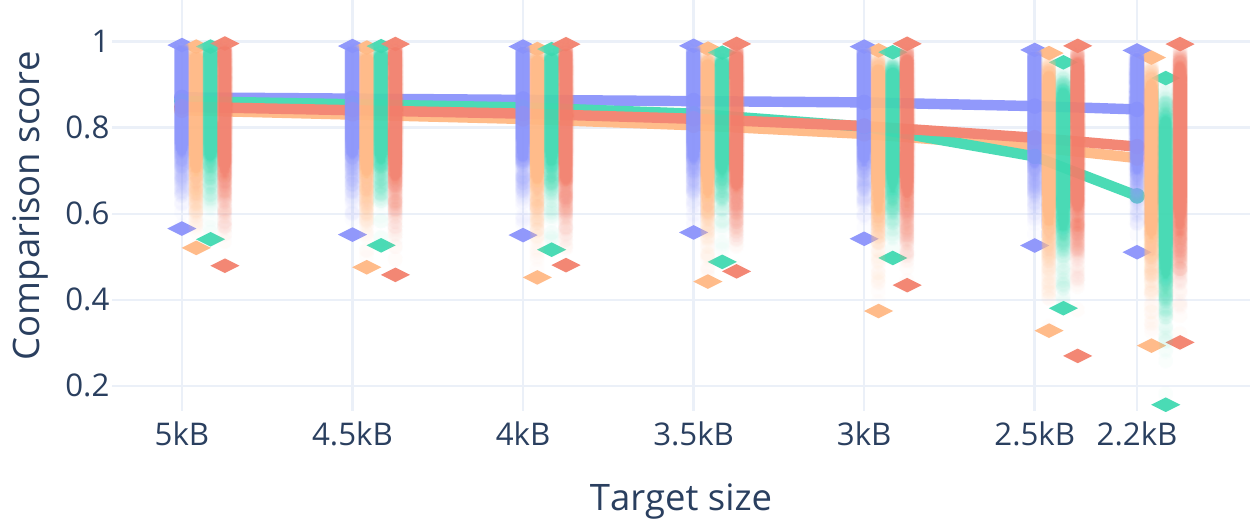}\\
Mated-self:\\
\includegraphics[width=\linewidth,valign=c]{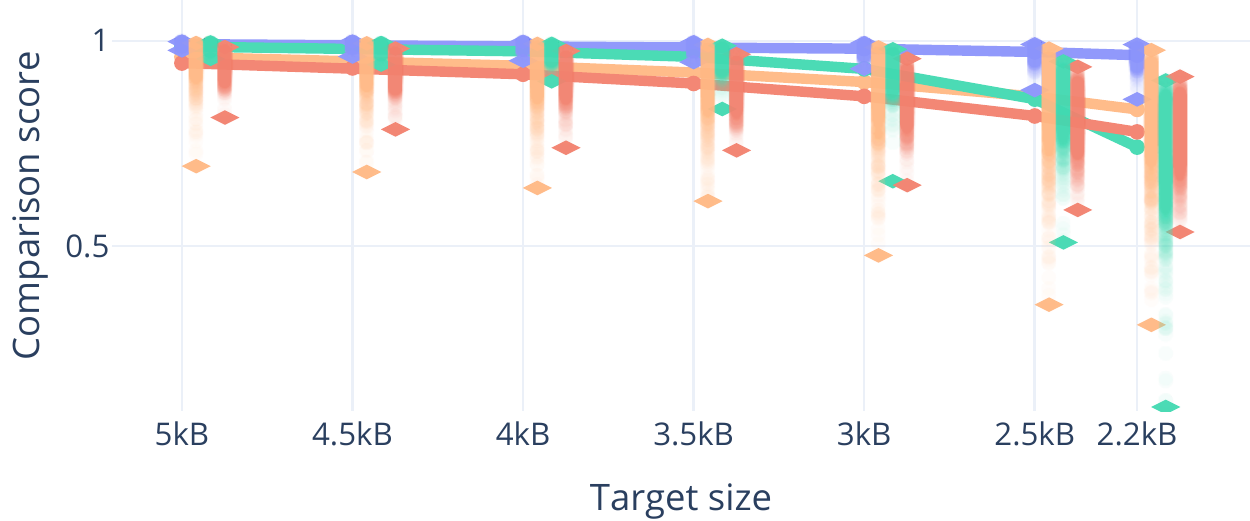}\\
\\\includegraphics[width=\linewidth,valign=c]{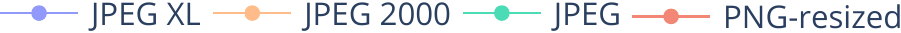}
\caption{\label{fig:cs-roi-mated} Comparison score results.}
\vspace{-1.5em}
\end{figure}

The comparison scores
shown in \autoref{fig:cs-roi-mated}
are similarity scores \cite{ISO-IEC-2382-37-2022} in the range $[-1,+1]$, meaning that higher values are intended to indicate higher facial similarity.
The y-axis shows the comparison score,
and the x-axis shows the target sizes in descending order from left to right.
The curves correspond to the different compression types.
Each linearly interpolated curve shows the mean value of the comparison scores per target size.
The transparent and thus ``smeared'' looking markers of the same color show all concrete individual comparison score data points per target size,
whereby the x-axis location is slightly shifted across the compression types for the sake of visibility.
Additional opaque rhombus-shaped markers highlight the maximum and minimum comparison scores.

In the results for the mated-other comparison trial configuration,
JPEG XL performed best both in terms of maintaining high mean comparison scores and in terms of the worst-case (minimum) comparison scores.
The compression types' results diverged most distinctly for the lowest target size, 2.2kB.
At this target size, JPEG's performance in terms of the mean score and in terms of the overall worst-case score dropped noticeably,
relative to slightly higher target sizes at which JPEG's results remained closer to or better than JPEG 2000 and PNG-resized.
A perhaps surprising result can be seen for PNG-resized,
which remained competitive with JPEG and JPEG 2000 across the target sizes,
and even yielded slightly better results than both at 2.2kB.

Results for the mated-self comparison trial configuration
lead to similar conclusions as for mated-other, with some differences.
JPEG XL is again the compression type with the best results, here with worst-case scores that are even more distinguished from those of the other compression types.
JPEG is again the worst performing compression at 2.2kB, while being competitive or better than JPEG 2000 and PNG-resized at higher target sizes.
Regarding differences to mated-other,
the order of the JPEG 2000 and PNG-resized mean score curves is reversed,
JPEG 2000 now being slightly better in this metric,
although PNG-resized now beats JPEG 2000 in terms of worst-case scores across all target sizes.

To summarize,
the results let us recommend JPEG XL across all target sizes,
while JPEG should be avoided for very low sizes (2.2kB),
but is otherwise competitive with JPEG 2000 and PNG-resized.
For the mated-self score spread and worst-case results,
the lead of the JPEG XL results was more pronounced even at higher shown target sizes,
with only JPEG showing competitive results at approximately 4kB and above.

\vspace{-0.5em}\subsection{Effect on quality scores}

\begin{figure}
\newcommand{\ingr}[1]{\includegraphics[width=0.48\linewidth,valign=c]{icn/#1}}
\centering
\setlength\tabcolsep{3pt}
\begin{tabular}{ll}
PNG-resized: & JPEG: \\
\ingr{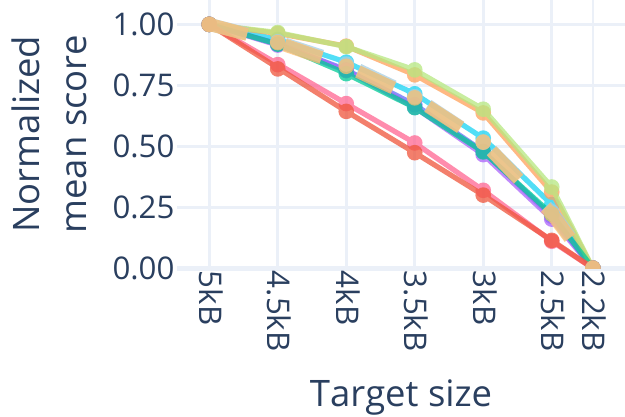} & \ingr{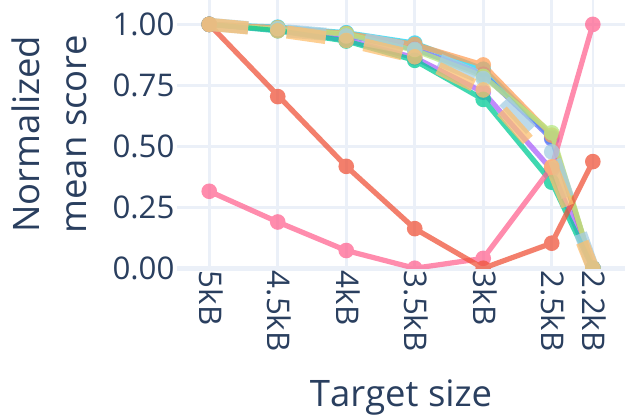} \\
JPEG 2000: & JPEG XL: \\
\ingr{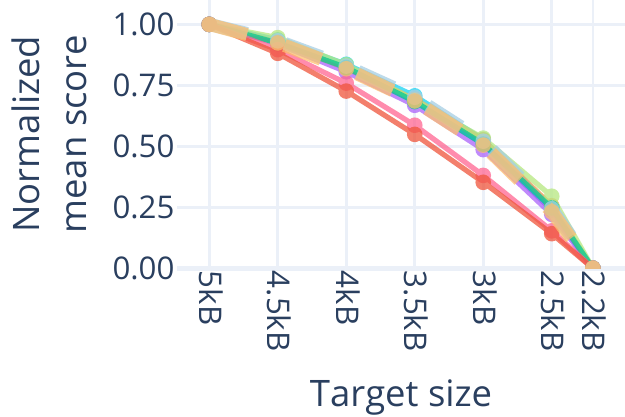} & \ingr{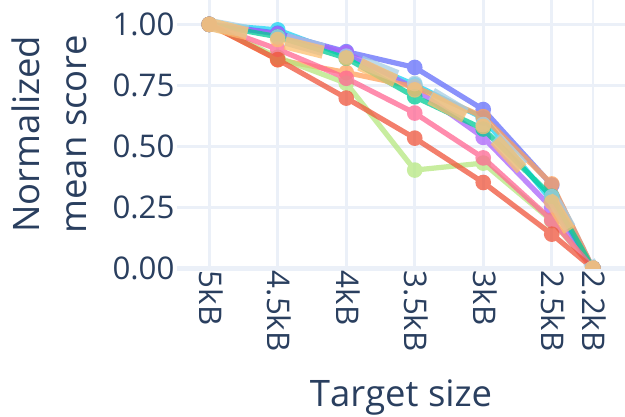} \\
\end{tabular}
\\[0.5em]\includegraphics[width=\linewidth,valign=c]{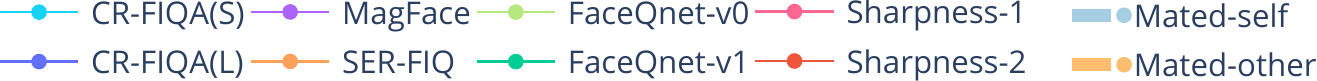}
\caption{\label{fig:norm-qs-cs-curves} Normalized mean quality and mated-self/other comparison score curves.}
\vspace{-1em}
\end{figure}

A direct computation of the correlation between the quality scores and the comparison scores is not possible,
since a quality score corresponds to one individual image, while a comparison score corresponds to an image pair.
Thus to analyze the extent to which the effect of the lossy image compression on the comparison scores
is reflected in the quality scores of the various tested FIQA methods,
we chose to examine the normalized mean score curves as shown in \autoref{fig:norm-qs-cs-curves}.
The x-axis shows the same target size range as for the comparison scores in \autoref{fig:cs-roi-mated},
while the y-axis now shows the mean quality or comparison scores,
normalized to $[0,1]$ since the original value ranges differed.
The plots  are separated by the four compression types,
and the various curves represent the eight FIQA methods as well as the comparison scores for the mated-self and mated-other trial configuration.
These normalized curves show that the effect of the lossy image compression types on most of the FIQA methods is similar to the effect on the comparison scores.
The two sharpness measures are the most clear exception,
especially for JPEG,
where the severe sharpness value increase for target size 2.5kB and 2.2kB
can be explained by the more severe JPEG ``block'' compression artifacts.
To better quantify how well each FIQA method captures the compression effect on the comparison scores,
we computed the distances between these normalized quality/comparison score curves,
shown in \autoref{tab:norm-qs-cs-distances}.
More specifically, we computed the areas under each linearly interpolated curve,
and then computed the absolute value of the difference between areas for each pairing of FIQA method and mated-other/mated-self comparison configuration.

Overall, the two sharpness metrics appeared distinctly less reliable than the six modern utility-centric FIQA models.
Of those six models, FaceQnet-v0 performed slightly worse overall, and substantially worse for JPEG XL.
FaceQnet-v1, which is architecturally identical to FaceQnet-v0, did however perform very well both overall and for JPEG XL,
which could indicate that the training procedure might be more important than the architectural differences of these FIQA models,
at least in the context of this compression effect scenario.
The differences between the results for the remaining FIQA models appear to be comparatively minor overall.

\begin{table}
\newcommand*\rot{\rotatebox{90}}
\newcommand{\bestv}[1]{\textbf{#1}}
\renewcommand{\arraystretch}{0.92}
\footnotesize \centering
\begin{tabular}{rrrrrrrrrr}
 &  & \rot{CR-FIQA(S)} & \rot{CR-FIQA(L)} & \rot{MagFace} & \rot{SER-FIQ} & \rot{FaceQnet-v0} & \rot{FaceQnet-v1} & \rot{Sharpness-1} & \rot{Sharpness-2} \\
\hline
\multirow{5}{*}{\rot{Mated-other}}
 & PNG-resized & \bestv{2} & \bestv{2} & \bestv{2} & 7 & 8 & \bestv{2} & 13 & 15 \\
 & JPEG & 5 & 4 & \bestv{0} & 5 & 4 & 2 & 69 & 43 \\
 & JPEG 2000 & 2 & \bestv{0} & 1 & 1 & 2 & \bestv{0} & 7 & 9 \\
 & JPEG XL & 2 & 5 & 2 & 4 & 13 & \bestv{1} & 7 & 14 \\
 & Combined & 3 & 3 & \bestv{1} & 4 & 7 & \bestv{1} & 24 & 20 \\
\hline
\multirow{5}{*}{\rot{Mated-self}}
 & PNG-resized & \bestv{1} & 2 & 3 & 7 & 8 & 3 & 13 & 15 \\
 & JPEG & 2 & \bestv{1} & 3 & 3 & \bestv{1} & 5 & 71 & 45 \\
 & JPEG 2000 & \bestv{0} & 1 & 2 & 1 & 1 & 1 & 8 & 10 \\
 & JPEG XL & \bestv{1} & 4 & 2 & 4 & 14 & \bestv{1} & 9 & 15 \\
 & Combined & \bestv{1} & 2 & 3 & 4 & 6 & 2 & 25 & 21 \\
\end{tabular}
\vspace{-0.5em}
\caption{\label{tab:norm-qs-cs-distances} Distances between normalized quality and mated-other/self comparison score curves from \autoref{fig:norm-qs-cs-curves}, as rounded $[0,100]$ percentage values (lower is better). For each row the lowest values are printed in \bestv{bold}. The ``combined'' rows show the mean values across the four compression types.}
\vspace{-1em}
\end{table}

\vspace{-0.5em}\subsection{Additional results}\vspace{-0.5em}

Besides the primary results presented in the previous subsections,
we also examined higher target sizes for the ROI images (27kB, 22kB, 17kB, 12kB, 10kB),
evaluated the same experiments with a second ``portrait'' image variant (with target sizes 30kB, 24kB, 18kB, 12kB, 10kB, 9kB, 8kB, 7.7kB),
and analyzed a non-mated comparison trial configuration.

For the portrait image variant the ColorFERET images were cropped to the minimal allowed width and height relative to the Inter-Eye Distance (IED) and Eye-Mouth Distance (EMD) as required
by ISO/IEC 39794-5:2019 \cite{ISO-IEC-39794-5-2019}.
The cropped portrait region was out of the original image bounds for 43 images,
with the out of bounds pixels being filled with black.
Other standard requirements were not checked and the IED was below the 90 pixel minimum for 243 images,
while an additional 751 images fell below the 120 pixel IED ``best practice'' value of the standard,
so images in this portrait variant were only partially compliant.
The comparison score results did however not differ substantially between the compression types for the portrait image variant,
nor with the higher target sizes for the ROI image variant.

The ``non-mated'' comparison trial configuration consisted out of randomly selected non-mated \cite{ISO-IEC-2382-37-2022} trials,
equal in number to the mated-other trials (3484),
likewise comparing images with the same target size compression applied.
Results for this configuration on the ROI images were arguably less interesting for the analysis,
due to only comparatively minor changes in the mean comparison scores even at the lower target sizes,
with substantial overlap of the computed individual scores and their minima/maxima.
We can however note that PNG-resized was comparatively most influential in terms of altering the mean score and the worst-case (maximum) scores,
which may indicate that the seemingly competitive mated configuration results for PNG-resized were facilitated in part
by removing or ``blurring'' more recognition-relevant information
in a similar manner across images.
Nevertheless, the differences were minor, and so PNG-resized appeared to be competitive with JPEG and JPEG 2000 given the results overall, especially at higher target sizes.

\vspace{-0.5em}\section{Conclusion}\vspace{-0.5em}
\label{sec:conclusion}

Based on the results we recommend JPEG XL for face image compression across all target sizes,
both for face ROI images and portrait images.
The differences in the effect on face recognition between compression types did become substantial for face ROI compression at sizes approximately below 5kB.
For higher sizes, above 10kB, JPEG XL can still provide the best worst-case results,
but the other tested compression types appeared viable at these sizes as well,
including the unusual PNG-resized compression.
While JPEG compression for ROI images led to slightly better results than JPEG 2000 and PNG-resized for target sizes at and above 3.5kB,
it also led to mostly worse results for the very low target sizes 3kB, 2.5kB, and 2.2kB.
The advantage of JPEG 2000 over JPEG at lower target sizes supports prior work results mentioned in the introduction \cite{Funk-QualityPerformance-IEEE-WestPoint-2005}\cite{Delac-ImageCompressionEffectsFaceRecognition-2007}.

Regarding FIQA,
the two sharpness measures did not capture the effect of compression on the comparison scores well, especially not for JPEG,
while the six modern utility-centric FIQA models mostly did.
If only the effect of the recommended JPEG XL compression on the utility were to be measured,
then the results indicate FaceQnet-v1, CR-FIQA(S), and MagFace as the primary candidates.
To more specifically select one of the six FIQA models for general use,
we do however recommend to refer to more general FIQA performance evaluations.

Future work could investigate the qualitative optimization of compression settings,
or evaluate the compression speed and assess the
potential for computational performance optimizations.
Future development work could furthermore attempt to create specialized face image compression types.

\clearpage

\bibliographystyle{IEEEbib}
\bibliography{refs}

\end{document}